# Joint Subspace Recovery and Enhanced Locality Driven Robust Flexible Discriminative Dictionary Learning

Zhao Zhang, *Senior Member*, *IEEE*, Jiahuan Ren, Weiming Jiang, Zheng Zhang, Richang Hong, *Member*, *IEEE*, Shuicheng Yan, *Fellow*, *IEEE* and Meng Wang, *Senior Member*, *IEEE*

*Abstract* — We propose a joint subspace recovery and enhanced locality based robust flexible label consistent dictionary learning method called *Robust Flexible Discriminative Dictionary Learning* (RFDDL). RFDDL mainly improves the data representation and classification abilities by enhancing the robust property to sparse errors and encoding the locality, reconstruction error and label consistency more accurately. First, for the robustness to noise and sparse errors in data and atoms, RFDDL aims at recovering the underlying clean data and clean atom subspaces jointly, and then performs DL and encodes the locality in the recovered subspaces. Second, to enable the data sampled from a nonlinear manifold to be handled potentially and obtain the accurate reconstruction by avoiding the overfitting, RFDDL minimizes the reconstruction error in a flexible manner. Third, to encode the label consistency accurately, RFDDL involves a discriminative flexible sparse code error to encourage the coefficients to be soft. Fourth, to encode the locality well, RFDDL defines the Laplacian matrix over recovered atoms, includes label information of atoms in terms of intra-class compactness and inter-class separation, and associates with group sparse codes and classifier to obtain the accurate discriminative locality-constrained coefficients and classifier. Extensive results on public databases show the effectiveness of our RFDDL.

## I. Introduction

SPARSE Representation (SR) using dictionary learning (DL) has been playing an important role for image representation and recognition due to its great success to the image restoration [12-13], denoising [3] and classification [15], [18], [27]. To be specific, SR approximates data by a linear combination of a few compact items from a dictionary to minimize the reconstruction error [9], [22]. The superiority and properties of the dictionary is crucial to the success of representation learning. To use SR for classification, *SR based Classification* (SRC) algorithm [20] was proposed, which used entire training set as dictionary for representation and obtained the impressive recognition results. But computing the codes from the whole original training set suffers from the issues of obtaining the inaccurate coefficients caused by the noise and errors in original data, and inefficiency due to a large dictionary size, which may restrict its real-world applications. To solve these issues, many compact DL methods were proposed recently [8-9], [16-17], [22], [26], [28-35].

Based on the use of supervised prior information, existing compact algorithms can be roughly divided into unsupervised and discriminative ones. Unsupervised models mainly compute the dictionaries that are suitable for representing data without using any class information of samples, such as [1], [15], [20], of which *K-Singular Value Decomposition* (KSVD) [1] is one most popular model. KSVD learns an over-complete dictionary from training set to represent data reliably, but it is not suitable for classification. Another one popular unsupervised algorithm is *Sparse Dimensionality Reduction* (SDR) [24] that improves the result by jointly learning a projection to reduce unfavorable features and redundant information to some extent.

In contrast to unsupervised models, discriminative methods explicitly use the label information of training data to obtain discriminant dictionary and also improve classification [5], [9], [25], [31-34]. Existing discriminative methods can be further divided into two categories. The first category aims at learning category-specific or multiple dictionaries to promote inter-class discrimination, e.g., *DL with Structured Incoherence* (DLSI) [35], *DL with Commonality and Particularity* (COPAR) [42], *Fisher Discrimination DL* (FDDL) [22], *Joint Discriminative Dimensionality Reduction and DL* (JDDR-DL) [23], *Low-rank Shared DL* (LRSDL) [43], *Projective Dictionary Pair Learning* (DPL) [5], and *Structured Analysis Discriminative DL* (ADDL) [32]. The other kind of discriminative DL method is to compute a shared dictionary for all classes of samples and incorporate certain discriminant information by regularization to ensure the discriminating ability of sparse codes or dictionary, for instance *Discriminative KSVD* (D-KSVD) [25], *Label Consistent KSVD* (LC-KSVD) [9], *Locality Constrained and Label Embedding DL* (LCLE-DL) [33] and *Support Vector Guided DL* (SVGDL) [34]. LCLE-DL clearly combines the locality-based and label-based embedding for representation and classification.

Although enhanced representation and classification results have been gained by seeking compact overall or class-specific dictionaries, aforementioned discriminative DL algorithms still suffer from several drawbacks. First, the dictionary and sparse codes are usually computed in the original input space, but the included noise and sparse errors may directly degrade the data representation and classification tasks. Although certain efforts have been made, e.g., combining the dimensionality reduction with DL [23], [24] to jointly calculate the feature subspace of

- Zhao Zhang, R. Hong and M. Wang are now with the Key Laboratory of Knowledge Engineering with Big Data (Hefei University of Technology), Ministry of Education, China; also with School of Computer and Information, Hefei University of Technology, Hefei 230601, China. (e-mails: cszzhang@gmail.com, hongrc.hfut@gmail.com, eric.mengwang@gmail.com
- J. Ren and W. Jiang are with School of Computer Science and Technology, Soochow University, Suzhou, China. (e-mails: hmzry10086@outlook.com, cswmjiang@gmail.com)
- Zheng Zhang is with the Bio-Computing Research Center, Harbin Institute of Technology, Shenzhen, China, and also with the University of Queensland Brisbane, QLD 4072, Australia. (e-mail: darrenzz219@gmail.com)
- S. Yan is with Department of Electrical and Computer Engineering, National University of Singapore, Singapore 117583. (e-mail: eleyans@nus.edu.sg)

data to potentially reduce unfavorable features and redundant information, it usually fails to recover the sparse errors in data. Second, to encourage intra-class samples to deliver the similar sparse codes and those from different classes to have different ones, both LC-KSVD and LCLE-DL have used two different strategies. Specifically, LC-KSVD pre-defines a discriminative sparse codes matrix $Q$ [9], but setting all nonzero entries to ones is too hard, since it treats each nonzero entry equally and does not consider any local information that should be encoded in sparse codes [33]. Thus, proposing an effective flexible way to encode the discriminative sparse code error with a soft measure is desired. In contrast, LCLE-DL constructs a Gaussian kernel function based Laplacian matrix over learned atoms instead of training data to model the locality [33], but finding an optimal kernel width $\delta$ is never easy in reality. Third, existing methods usually seek the compact dictionary and codes by minimizing a reconstruction error between data and its linear reconstruction directly, which may suffer from the overfitting issue when the number of training data is limited. Also, the data sampled from a nonlinear manifold cannot be potentially handled. Thus, it will be better to derive a relaxed reconstruction so that the data sampled from nonlinear manifold can also be processed.

In this paper, we mainly propose the effective strategies to overcome aforementioned issues, and propose a robust flexible label consistent DL method to enhance the representation and classification abilities. The main contributions are shown as:

(1) A joint subspace recovery based enhanced discriminative locality constrained *Robust Flexible Discriminative Dictionary Learning* (RFDDL) model is proposed for joint representation and classification. RFDDL extends regular label consistent DL into the enhanced locality based robust flexible label consistent DL. Specifically, our RFDDL improves the representation and classification results by enhancing the robust properties of the discriminative DL procedures to noise and sparse errors, and encoding the discriminative locality over the recovered atoms, reconstruction error and label consistency more accurately. The classification error is also incorporated for joint optimization. RFDDL also abandons costly $L_0/L_1$-norm and use time-saving Frobenius-norm on coefficients for the group SR and efficiency. The relationship analysis with related work show that RFDDL is a more general, robust and powerful DL framework.

(2) For the robust properties of the DL process to noise and errors, RFDDL proposes to recover the underlying clean data and atom subspaces explicitly, and performs the discriminative DL in the recovered subspaces for robust data representations. To encourage samples of one class to have similar sparse codes and those of different classes to have different ones, and avoid the tricky issue of selecting optimal kernel width $\delta$ suffered in LCLE-DL, RFDDL encodes the locality and defines the graph Laplacian matrix by calculating the discrimination-promoting reconstruction weights based on recovered clean atoms rather than the original dictionary. For the discrimination-promoting learning, RFDDL includes the label information of atoms of the recovered dictionary in terms of high inter-class separation and intra-class compactness, and associates the Laplacian matrix with the coefficients and classifier learning to produce more accurate coefficients and discriminant classifier jointly. Thus, the local neighbours of each atom can be picked from the same class as much as possible for accurate similarity measure.

(3) To encode the sparse reconstruction error, discriminative sparse codes error and classification error more accurately, our RFDDL propose flexibly-relaxed extensions of them for more accurate data representation and classification. The flexible sparse reconstruction error over the recovered clean data can potentially enable the data sampled from a nonlinear manifold to be handled using the robust flexible label consistent DL. The discriminative flexible sparse code error encodes the mismatch among the discriminative sparse code matrix, enhanced locality constrained coefficients and a residue, which can provide a soft and flexible measure on the coefficients adaptively and can also address the hard constraint issue suffered in the LC-KSVD. The flexible classification error can enable the label fitness error to avoid the overfitting and make the prediction task accurately.

By unifying the joint subspace recovery on data and atoms, enhanced locality based flexible reconstruction error, flexible label consistency and flexible classification error, our RFDDL can obtain a robust compact discriminative dictionary and a set of structure preserving coefficients for data represenation.

The paper is outlined as follows. In Section II, we review the related work briefly. In Sections III, we present the formulation and optimization of RFDDL. Section IV shows the connections with other methods. Section V shows the experimental settings and results. Finally, the paper is concluded in Section VII.

## II. RELATED WORK

### A. Review of D-KSVD and LC-KSVD

We briefly review the D-KSVD and LC-KSVD formulations.

**D-KSVD.** To improve the classification results, D-KSVD combines a classification error into KSVD to learn a dictionary and a classifier jointly. Given a training set $X = [x_1, \cdots x_N] \in \mathbb{R}^{n \times N}$ containing $N$ samples from $c$ classes, where $x_i$ corresponds to an $n$-dimensional sample in original space, the joint dictionary and classifier learning problem of D-KSVD is defined as

$$\min_{D,S,W} \|X - DS\|_F^2 + \gamma \|H - WS\|_F^2 = \left\| \begin{bmatrix} X \\ \sqrt{\gamma}H \end{bmatrix} - \begin{bmatrix} D \\ \sqrt{\gamma}W \end{bmatrix} S \right\|_F^2, \quad (1).$$
$$s.t. \|s_i\|_0 \leq T_0, i = 1, 2, ..., N$$

where $D = [d_1 \cdots d_K] \in \mathbb{R}^{n \times K}$ is the dictionary containing $K$ items, $S = [s_1, \cdots, s_N] \in \mathbb{R}^{K \times N}$ are the coefficients of $X$, $\|s_i\|_0$ counts the number of nonzero elements in $s_i$, $W = [w_1, ..., w_K] \in \mathbb{R}^{c \times K}$ is the classifier, and $T_0$ is a sparse constraint factor to ensure that the coefficient vector of each $x_i$ have fewer than $T_0$ nonzero items. $H = [h_1, h_2, \cdots h_N] \in \mathbb{R}^{c \times N}$ is the label set of all the samples, where $h_i = [0, 0 \cdots 1 \cdots 0, 0]^T \in \mathbb{R}^c$ is the label vector of data point $x_i$ with the nonzero position indicating its class assignment.

**LC-KSVD.** Based on the modeling of D-KSVD, LC-KSVD further includes a discriminative sparse-code error to enforce the label consistency and encourage the structures of the codes to be preserved so that each data can be reconstructed by those from a class as much as possible. The criterion of LC-KSVD is

$$\min_{D,S,A,W} \|X - DS\|_F^2 + \alpha \|Q - AS\|_F^2 + \beta \|H - WS\|_F^2$$
$$= \left\| \begin{bmatrix} X \\ \sqrt{\alpha}Q \\ \sqrt{\beta}H \end{bmatrix} - \begin{bmatrix} D \\ \sqrt{\alpha}A \\ \sqrt{\beta}W \end{bmatrix} S \right\|_F^2, \quad s.t. \|s_i\|_0 \leq T_0, i = 1, 2, ..., N \quad (2)$$

where $H$ is the label set with binary entries (0 or 1), $\alpha$ and $\beta$ are positive scalars. $Q=[q_1,q_2,...,q_N] \in \mathbb{R}^{K \times N}$ are "*discriminative*" sparse codes for $X$, where $q_i = [0\cdots,1,1,\cdots 0]^T \in \mathbb{R}^K$ is an ideal "discriminative" sparse code for $x_i$ if the nonzero values of $q_i$ occur at those indices where $x_i$ and atom $d_i$ share the same label [9]. Suppose that $X=\{x_i\}_{i=1}^9$ has 3 classes, where $x_1, x_2, x_3, x_4$ are from class 1, $x_5, x_6$ are from class 2, and the rest from class 3, the "discriminative" sparse codes matrix $Q$ is defined as

$$Q = \begin{bmatrix} 1 & 1 & 1 & 1 & 0 & 0 & 0 & 0 & 0 \\ 1 & 1 & 1 & 1 & 0 & 0 & 0 & 0 & 0 \\ 0 & 0 & 0 & 0 & 1 & 1 & 0 & 0 & 0 \\ 0 & 0 & 0 & 0 & 1 & 1 & 0 & 0 & 0 \\ 0 & 0 & 0 & 0 & 0 & 0 & 1 & 1 & 1 \\ 0 & 0 & 0 & 0 & 0 & 0 & 1 & 1 & 1 \end{bmatrix}. \quad (3)$$

Thus, term $\|Q-AS\|_F^2$ is the discriminative sparse-code error, where $A \in \mathbb{R}^{K \times K}$ converts the original sparse codes in $S$ to be more discriminative in a feature subspace $\mathbb{R}^K$. $\|Q-AS\|_F^2$ can encourage the label consistency in the resulting codes, but $A$ is arbitrarily defined, so it cannot preserve local information and inherit the structure information of samples. Setting all nonzero entries to ones is also too hard, since the coefficients in $S$ are essentially soft, i.e., a large value $s_{i,j}$ means that the contribution of each $x_j$ to reconstruct $x_i$ is large, and small otherwise.

Note that D-KSVD and LC-KSVD can be equivalent with the uniform atom allocation [36], which is also based on the identical initialization conditions and optimization methods. By the equivalence, we can reformulate LC-KSVD as

$$\langle D^*, S^*, W^* \rangle \overset{\gamma=\upsilon\alpha+\beta}{=} \arg\min_{D,S,W} \left\| \begin{bmatrix} X \\ \sqrt{\gamma}H \end{bmatrix} - \begin{bmatrix} D \\ \sqrt{\gamma}W \end{bmatrix} S \right\|_F^2, \quad (4)$$

$$s.t. \|s_i\|_0 \le T_0, i=1,2,...,N$$

which is just the problem of D-KSVD, where $\upsilon$ is the number of dictionary atoms allocated per class. The analysis in [36] also shows that D-KSVD is preferable because of its simplicity and efficiency, compared to the LC-KSVD algorithm.

### B. Review of LCLE-DL

LCLE-DL is another one related DL method, so we also briefly revisit it. LCLE-DL calculates a discriminative dictionary $D$ by combining the label embedding of atoms and locality constraint of atoms jointly. The problem of LCLE-DL is defined as

$$\min_{D,S,V,L} \|X-DS\|_2^2 + \alpha tr(S^T LS) + \|X-DV\|_2^2 + \beta tr(V^T UV) + \gamma \|S-V\|_2^2, \quad s.t. \|d_i\|^2 = 1, \ i=1,...,K \quad (5)$$

where $S$ and $V$ denote the coefficient matrices, $\|X-DS\|_2^2$ and $\|X-DV\|_2^2$ denote the reconstruction error, and $\|S-V\|_2^2$ is the regularization used to transfer the label constraint $Tr(V^T UV)$ to/from the locality constraint $Tr(S^T LS)$. $U$ is the scaled label matrix constructed by the labels of dictionary. $\alpha, \beta$ and $\gamma$ are parameters. $L$ is a graph Laplacian matrix defined as

$$L=G-M, \text{ where } G=diag(g_1,...g_K), \ g_i=\sum_{j=1}^K M_{i,j}, \quad (6)$$

where the nearest neighbor graph is weighted by $M$ as

$$M_{i,j} = \begin{cases} \exp(-\|d_i-d_j\|_2/\delta) & \text{if } d_j \in kNN(d_i) \\ 0 & \text{else} \end{cases}, \quad (7)$$

which is defined by the Gaussian function, where $\delta$ is the kernel width, $kNN(d_i)$ is the k-nearest neighbor set of atom $d_i$, $G$ is a diagonal matrix and $M_{i,j}$ encodes the similarity between atoms $d_i$ and $d_j$. By the above definitions, $\|X-DS\|_2^2 + \alpha Tr(S^T LS)$ can encode the reconstruction error with the locality constraint, where $Tr(S^T LS)$ inherits the manifold structure of training set. $\|X-DV\|_2^2 + \beta Tr(V^T UV)$ encodes the reconstruction error with label embedding, where $Tr(V^T UV)$ forces the intra-class atoms in $D$ to have similar profiles. Term $\|S-V\|_2^2$ is a regularization over the coefficients, which ensures the mutual transformation between the label embedding and locality constraint.

### III. ROBUST FLEXIBLE DISCRIMINATIVE DICTIONARY LEARNING (RFDDL)

### A. Problem Formulation

The presented RFDDL model improves the robust properties of discriminative DL to noise and sparse errors in twofold. First, it calculates the robust discriminative dictionary and codes in the recovered clean data and atom subspaces. Specifically, RFDDL decomposes the original $X$ and dictionary $D$ in each iteration to recover underlying clean data $X_{new}$ and clean dictionary $D_{new}$, and models the errors $E$ and $E_D$ at the same time in terms of $X = X_{new} + E$ and $D = D_{new} + E_D$, where $L_{2,1}$-norm is used on $E$ and $E_D$ so that the sparse errors in data and atoms can be corrected jointly. Then, RFDDL performs the discriminant DL over clean $X_{new}$ and $D_{new}$ for accurate representations. Second, our RFDDL encodes the locality and defines a Laplacian matrix by computing discrimination-promoting reconstruction weights over recovered clean atoms, which can encourage intra-class samples to have similar sparse codes and inter-class samples to have different ones, and will be detailed in next subsection. By combing the joint subspace recovery and Laplacian regularized reconstruction error, discriminative sparse-code error and data classification error, the initial problem of RFDDL is given as

$$\min_{D,S,L,E,E_D,W} \|X_{new} - D_{new}LS\|_F^2 + \alpha \left(\|E^T\|_{2,1} + \|E_D^T\|_{2,1}\right) + \beta \left(\|Q-LS\|_F^2 + \|SH_e\|_F^2\right) + \gamma \left(\|H-WLS\|_F^2 + \|W^T\|_{2,1}\right), \quad (8)$$

$$s.t. \ X = X_{new} + E, \ D = D_{new} + E_D$$

where $\alpha \left(\|E^T\|_{2,1} + \|E_D^T\|_{2,1}\right)$ is the $L_{2,1}$-norm based error, $\alpha, \beta$ and $\gamma$ are the parameters. Note that the $L_{2,1}$-norm can ensure the regularized matrix to be sparse in rows, and the $L_{2,1}$-norm based metric is robust to noise and outliners in data and atoms [2], [7], [27]. $L_{2,1}$-norm based classifier $\|W^T\|_{2,1}$ can force the columns of $W$ to be sparse so that the discriminative soft labels can be predicted in the latent sparse subspace. $Q$ and $H$ are similarly defined as LC-KSVD, and $\|SH_e\|_F^2$ is the Frobenius-norm based coefficients, $H_e = I - ee^T/N$ is the "centering matrix", that is, $SH_e = S - See^T/N$ can be considered as the normalized coding coefficients. It is clear that the discriminative Laplacian matrix is associated with the learning of codes and classifier, which can potentially obtain the more accurate codes, discriminative dictionary and powerful discriminative classifier jointly.

Please note that the linear reconstruction $\|X_{new} - D_{new}LS\|_F^2$ may be overfitted especially when the number of training data is limited. To enable the data sampled from nonlinear manifold to be handled potentially by DL and avoid the overfitting, our RFDDL proposes a flexible reconstruction residue motivated by [38], [39]. Note that the discriminative sparse code error $\|Q - LS\|_F^2$ in above problem is different from that of LC-KSVD, since $L$ is explicitly defined as a Laplacian matrix in RFDDL rather than a random matrix as the LC-KSVD. To address the suffered hard constraint issue when minimizing the mismatch between $Q$ and $LS$ directly, RFDDL defines a discriminative flexible sparse code error. The flexible reconstruction residue and discriminative flexible sparse code error are defined as

$$\left\|X_{new} + b_1 e^T - D_{new} S\right\|_F^2 \text{ and } \left\|Q + b_2 e^T - LS\right\|_F^2, \quad (9)$$

where $b_1 \in \mathbb{R}^{n \times 1}$, $b_2 \in \mathbb{R}^{N \times 1}$ are bias and $e \in \mathbb{R}^{N \times 1}$ is column vector of all ones. That is, $\left\|X_{new} + b_1 e^T - D_{new} S\right\|_F^2$ encodes the mismatch between $X_{new} + b_1 e^T$ and $D_{new} S$, and $\left\|Q + b_2 e^T - LS\right\|_F^2$ encodes the mismatch between $Q + b_2 e^T$ and $LS$ rather than between $Q$ and $LS$ directly. Note that the flexible reconstruction error and discriminative flexible sparse code error in our RFDDL clearly differs from [38], [39] that have discussed label prediction for classification. Also, RFDDL defines the flexible reconstruction in recovered clean space rather than original space. In addition, the purpose of defining $\left\|Q + b_2 e^T - LS\right\|_F^2$ is to keep the structures of the coefficients, which is also obviously different.

By combing the joint subspace recovery, enhanced locality based reconstruction error $\left\|X_{new} + b_1 e^T - D_{new} S\right\|_F^2$, discriminative flexible sparse code error $\left\|Q + b_2 e^T - LS\right\|_F^2$ and the classification error, the problem of our RFDDL can be reformulated as

$$\begin{aligned}\min_{\substack{D,S,L,b_1,b_2,\\b_3,E,E_D,W}} &\left\|X_{new} + b_1 e^T - D_{new} LS\right\|_F^2 + \alpha\left(\left\|E^T\right\|_{2,1} + \left\|E_D^T\right\|_{2,1}\right) \\ &+ \beta\left(\left\|Q + b_2 e^T - LS\right\|_F^2 + \left\|SH_e\right\|_F^2\right) \\ &+ \gamma\left(\left\|H + b_3 e^T - WLS\right\|_F^2 + \left\|W^T\right\|_{2,1}\right)\end{aligned}, \quad (10)$$

$$s.t. \; X = X_{new} + E, \; D = D_{new} + E_D$$

where $\left\|H + b_3 e^T - WLS\right\|_F^2$ is the classification error, which is also defined in a flexible manner to facilitate the optimization and can also avoid the possible overfitting issue in the label fitness measure and hence make the prediction results more accurately. $WLs_i$ is the predicted soft label vector of $x_i$, where the biggest entry in $WLs_i$ (ideally, 1) decides the label of $x_i$. But note that the biggest entry in $WLs_i$ is not necessarily strictly 1 in reality, since the hard label is the same as long as the position according to the biggest entry of $WLs_i$ is correct.

By substituting the errors $E$ and $E_D$ back into the problem of Eq.(10), we can have the following final problem:

$$\begin{aligned}\min_{\substack{D,S,L,b_1,b_2,\\b_3,E,E_D,W}} &\left\|(X - E) + b_1 e^T - (D - E_D) LS\right\|_F^2 + \alpha\left(\left\|E^T\right\|_{2,1} + \left\|E_D^T\right\|_{2,1}\right) \\ &+ \beta\left(\left\|Q + b_2 e^T - LS\right\|_F^2 + \left\|SH_e\right\|_F^2\right) \\ &+ \gamma\left(\left\|H + b_3 e^T - WLS\right\|_F^2 + \left\|W^T\right\|_{2,1}\right)\end{aligned}. \quad (11)$$

### B. Discriminative Graph Laplacian Matrix Construction

To encode the neighborhood of samples more accurately and avoid the tricky issue of selecting the optimal kernel width, our RFDDL defines the graph Laplacian matrix $L$ based on the enhanced discriminative locality-based reconstruction weight matrix $M$. Specifically, we incorporate label information of the atoms of $D_{new}$ into the construction of the weight matrix $M$. To fully use supervised the class information of atoms, we use a similar idea as [40] to refine the distances between the atoms so that the resulted neighborhood of data is more discriminating and accurate. That is, we increase the pre-calculated distances between those inter-class atoms and reduce the distances of those intra-class atoms artificially as

$$(\Delta_{new})_{i,j} = \begin{cases} \Delta_{i,j} + \max(\Delta_{i,j}), & \text{if } l(d_i^{new}) \neq l(d_j^{new}) \\ \dfrac{\Delta_{i,j}}{\max(\Delta_{i,j})}, & \text{else if } l(d_i^{new}) = l(d_j^{new}) \end{cases}, \quad (12)$$

where $d_i^{new}$ is the $i$-th atom of recovered clean dictionary $D_{new}$, $\Delta$ is the original Euclidean distance matrix, $\max(\Delta_{i,j})$ is the largest distance between all pairs of atoms in $\Delta$, and $\Delta_{new}$ is the newly-defined distance matrix. It is clear from $\Delta_{new}$ that the distances between intra-class atoms can be reduced so that the local neighbors of each atom can be picked from the same class as much as possible. Based on the new distance matrix $\Delta_{new}$, the reconstruction weight matrix $M$ can be computed by

$$\varepsilon(M) = \sum_i \left\| d_i^{new} - \sum_{j: d_j^{new} \in kNN(d_i^{new})} M_{i,j} d_j^{new} \right\|^2, \quad (13)$$

$$s.t. \sum_{j: d_j^{new} \in kNN(d_i^{new})} M_{i,j} = 1, \; M_{i,j} \geq 0$$

where $kNN(d_i^{new})$ is the $k$-nearest neighbor set of the atom $d_i^{new}$ over $\Delta_{new}$. By repeating the above step for each atom, we can obtain $M = [M_{i,j}] \in \mathbb{R}^{K \times K}$. After calculating $M$, we symmetrize it as $M \leftarrow (M + M^T)/2$ and then normalize it as $M = D^{-1/2} M D^{-1/2}$, where $D$ is a diagonal matrix with entries $D_{ii} = \sum_j M_{i,j}$. By the normalized $M$, the Laplacian matrix $L$ can be similarly obtained as $L = G - M$, where $G$ is a diagonal matrix with entries being $g_{ii} = \sum_j M_{i,j}$. As a result, although the Frobenius-norm is used, the codes may still be accurate by considering local information of dictionary and designing enhanced graph Laplacian matrix. Also, minimizing $\left\|Q + b_2 e^T - LS\right\|_F^2$ can force intra-class samples to have very similar atoms and representation, i.e., encouraging the label consistency in the resulting atoms and sparse codes.

In what follows, we mainly describe the optimization and convergence of our proposed RFDDL.

### C. Optimization

We show the optimization procedures of RFDDL. Let $\wp$ be the objective function of our RFDDL method in Eq.(11), by taking the derivative of $\wp$ w.r.t. bias $b_1$, $b_2$ and $b_3$, and setting the derivatives to zeros respectively, we can obtain

$$\frac{\partial \wp}{\partial b_1} = b_1 e^T e + X_{new} e - D_{new} LSe = 0 \Rightarrow b_1 = (D_{new} LSe - X_{new} e)/N, \quad (14)$$

$$\frac{\partial \wp}{\partial b_2} = b_2 e^T e + Qe - LSe = 0 \Rightarrow b_2 = (LSe - Qe)/N, \quad (15)$$

$$\frac{\partial \wp}{\partial b_3} = b_3 e^T e + He - WLSe = 0 \Rightarrow b_3 = (WLSe - He)/N. \quad (16)$$

Based on the above equations, we can rewrite the flexible reconstruction error $X_{new} + b_1 e^T - D_{new} LS$, discriminative flexible sparse codes error $Q + b_2 e^T - LS$ and the flexible classification error $H + b_3 e^T - WLS$ as follows:

$$\begin{aligned}&X_{new} + b_1 e^T - D_{new} LS \\ &= X_{new} + (D_{new} LSee^T - X_{new} ee^T)/N - D_{new} LS \\ &= X_{new} H_e - D_{new} LSH_e = (X - E)H_e - (D - E_D)LSH_e\end{aligned}, \quad (17)$$

$$Q + b_2 e^T - LS = Q + \left(LSee^T - Qee^T\right)/N - LS = QH_e - LSH_e, \quad (18)$$

$$H + b_3 e^T - WLS = H + \left(WLSee^T - Hee^T\right)/N - WLS$$
$$= (H - WLS)H_e \quad (19)$$

where $H_e = I - ee^T/N$ is the "centering matrix". By substituting the above equations into Eq.(11), we can obtain the following equivalent optimization problem for RFDDL:

$$\min_{D,S,L,E,E_D,W} \|(X-E)H_e - (D-E_D)LSH_e\|_F^2 + \alpha\left(\|E^T\|_{2,1} + \|E_D^T\|_{2,1}\right)$$
$$+ \beta\left(\|QH_e - LSH_e\|_F^2 + \|SH_e\|_F^2\right) \quad (20)$$
$$+ \gamma\left(\|HH_e - WLSH_e\|_F^2 + \|W^T\|_{2,1}\right)$$

From the above problem, it is easy to check that the variables depend on each other, so they cannot be solved directly. By following the common procedures, we solve the problem by updating them alternately. Firstly, we initialize the dictionary $D$ as a random matrix, and initialize $E$ and $E_D$ as zero matrices. Then, RFDDL can be optimized using the following steps:

**1) Robust Flexible Label Consistent Dictionary Learning and Clean Dictionary Atom Recovery:**
*Fix others, update $D$ and $S$.* We first fix the recovered clean data $X_{new}$, sparse errors $E_D$ and graph Laplacian matrix $L$ for the robust flexible label consistent dictionary learning:

$$\min_{D,S} \|X_{new}H_e - (D-E_D)LSH_e\|_F^2 + \beta\left(\|QH_e - LSH_e\|_F^2 + \|SH_e\|_F^2\right)$$
$$+ \gamma\|HH_e - WLSH_e\|_F^2 \quad (21)$$

By taking the derivative w.r.t. $D$ and zeroing the derivative, we can update the dictionary $D_{t+1}$ at the $(t+1)$-th iteration as

$$D_{t+1} = \left(\left(X_{new}\right)_t + \left(E_D\right)_t L_t S_t\right)H_e H_e^T S_t^T L_t^T \left(L_t S_t H_e H_e^T S_t^T L_t^T\right)^{-1}. \quad (22)$$

Similarly, by taking the derivative w.r.t. $S$ and setting it to 0, we can update the coefficients matrix $S_{t+1}$ accordingly as

$$S_{t+1} = \left(L_t^T (D_{new})_{t+1}^T (D_{new})_{t+1} L_t + \beta L_t^T L_t + \beta I + \gamma L_t^T W_t^T W_t L_t\right)^{-1}$$
$$\times \left(L_t^T (D_{new})_{t+1}^T (X_{new})_t + \beta L_t^T Q + \gamma L_t^T W_t^T H\right), \quad (23)$$

where $(D_{new})_{t+1} = D_{t+1} - (E_D)_t$ is the recovered clean dictionary and $(X_{new})_t = X - E_t$ is the recovered clean data matrix.

*Fix others, update $E_D$ and recover the clean dictionary $D_{new}$.* Given $D$ and $S$, we can update the sparse error $E_D$ in dictionary $D$ from the following reduced formulation:

$$\min_{E_D} \|(X-E)H_e - (D-E_D)LSH_e\|_F^2 + \alpha\|E_D^T\|_{2,1}. \quad (24)$$

According to the property of L$_{2,1}$-norm [2], [7], [19], [27], i.e., $E_D^T\|_{2,1} = 2tr(E_D \Lambda E_D^T)$, where $\Lambda$ is a diagonal matrix with the entries being $\upsilon_{ii} = 0.5/\|(E_D)_i\|_2$, $tr(\cdot)$ is trace operator and $(E_D)_i$ is the $i$-th column of sparse error $E_D$. When each $(E_D)_i \neq 0$, the above formulation can be easily approximated as

$$\min_{E_D,\Lambda} \|X_{new}H_e - (D-E_D)LSH_e\|_F^2 + \alpha\|E_D^T\|_{2,1}$$
$$= \|X_{new}H_e - (D-E_D)LSH_e\|_F^2 + \alpha \cdot tr(E_D \Lambda E_D^T), \quad (25)$$

By taking the derivative of the above problem w.r.t. $E_D$, we can update the error matrix $(E_D)_{t+1}$ as follows:

$$(E_D)_{t+1} = \left(D_{t+1}L_t S_t - (X_{new})_{t+1}\right)H_e H_e^T S_{t+1}^T L_t^T$$
$$\times \left(L_t S_{t+1}H_e H_e^T S_{t+1}^T L_t^T + \alpha\Lambda_t\right)^{-1}. \quad (26)$$

After obtaining the error matrix $(E_D)_{t+1}$, we can update the recovered dictionary $D_{new}$ as $(D_{new})_{t+1} = D_{t+1} - (E_D)_{t+1}$. After that, we can use $(D_{new})_{t+1}$ to update the graph Laplacian matrix $L_{t+1}$ by Eqs.(12) and (13), and update the diagonal matrix $\Lambda$ as

$$\Lambda_{t+1} = diag\left((\upsilon_{t+1})_{ii}\right), where\ (\upsilon_{t+1})_{ii} = 1/\left(2\|(E_D)_{t+1}^i\|_2\right). \quad (27)$$

**2) Recovering Underlying Clean Data Subspace:**
*Fix others, update the error $E$ and recover the clean data $X_{new}$.* We fix the recovered dictionary $D_{new}$ and coefficients matrix $S$ to encode the sparse error $E$ from the following problem:

$$\min_E \|(X-E)H_e - D_{new}LSH_e\|_F^2 + \alpha\|E^T\|_{2,1}. \quad (28)$$

By the definition of L$_{2,1}$-norm, we have $\|E^T\|_{2,1} = 2tr(EVE^T)$ similarly, where $V$ is a diagonal matrix with the entries being $v_{ii} = 0.5/\|e^i\|_2$, $e^i$ is the $i$-th column of $E$. Note that the above problem can be similarly approximated as the following one:

$$\min_{E,V} \|(X-E)H_e - D_{new}LSH_e\|_F^2 + \alpha \cdot tr(EVE^T). \quad (29)$$

when each $e^i \neq 0$, $i=1, 2,\ldots, N$. By taking the derivative of the above problem w.r.t. $E$, we can update the error $E_{t+1}$ as

$$E_{t+1} = \left(X - (D_{new})_{t+1}L_t S_{t+1}\right)H_e H_e^T \left(H_e H_e^T + \alpha V_t\right)^{-1}. \quad (30)$$

After $E_{t+1}$ is updated, the diagonal matrix $V$ is inferred as

$$V_{t+1} = diag\left((v_{t+1})_{ii}\right), where\ (v_{t+1})_{ii} = 1/\left(2\|e_{t+1}^i\|_2\right), \quad (31)$$

where $e^i$ is the $i$-th row of the error matrix $E_{t+1}$. Then, the clean data can be easily recovered as $(X_{new})_{t+1} = X - E_{t+1}$.

**3) Robust Discriminative Classifier Learning:**
*Fix others, update the linear classifier $W$.* We fix $S$ and use the updated Laplacian matrix $L$ to seek a robust linear multi-class classifier $W$. Since $\|W^T\|_{2,1} = 2tr(W\mathbb{Q}W^T)$, where $\mathbb{Q}$ is a diagonal matrix with the entries being $q_{ii} = 0.5/\|w_i\|_2$, we can similarly have the following approximated problem:

$$\min_{W,\mathbb{Q}} \gamma\left(\|HH_e - WLSH_e\|_F^2 + \|W^T\|_{2,1}\right)$$
$$= \gamma\left(tr\left((HH_e - WLSH_e)(HH_e - WLSH_e)^T\right) + tr(W\mathbb{Q}W^T)\right), \quad (32)$$

when each $W_i \neq 0$. By taking the derivative of the problem w.r.t. $W$, we can update the robust classifier $W_{t+1}$ as

$$W_{t+1} = HH_e H_e^T S_{t+1}^T L_{t+1}^T \left(L_{t+1}S_{t+1}H_e H_e^T S_{t+1}^T L_{t+1}^T + \mathbb{Q}_t\right)^{-1}. \quad (33)$$

After the linear classifier $W_{t+1}$ is updated at each iteration, the diagonal matrix $\mathbb{Q}$ can be inferred as

$$\mathbb{Q}_{t+1} = diag\left((q_{t+1})_{ii}\right), where\ (q_{t+1})_{ii} = 1/\left(2\|w_{t+1,i}\|_2\right), \quad (34)$$

where $w_{t+1,i}$ is the $i$-th column vector of $W_{t+1}$. For the complete representation of the optimization, we summarize the whole procedures in Algorithm 1, where the diagonal matrices $V$, $\Lambda$ and $\mathbb{Q}$ are all initialized to be the identity matrices, similarly as [2], [19] that have shown that this choice generally works well. The iteration stops when the difference between the objective function values in two adjacent iterations is less than 0.001.

## D. Classification Approach for Involving New Data

We show how to use RFDDL for representing and classifying outside new data. Since a linear label predictor $W^*$ is explicitly trained over the coefficients in training phase, we only need to obtain the coefficients $S_{test}$ of the test set $X_{test}$ and use $W^*S_{test}$ to predict the soft label matrix of $X_{test}$. Specifically, we propose two effective strategies by reconstruction and embedding to obtain the coefficient vector $s_{test}$ of each new data $x_{test}$ in $X_{test}$.

**Classification scheme one by reconstruction (RFDDL-r):** In this scheme, we compute the coding coefficients of $X_{test}$ by involving an efficient reconstruction process with well-trained clean dictionary $D^*_{new}$ for the representation learning:

$$\min_{S_{test}} \|X_{test}H_e - D_{new}S_{test}H_e\|_F^2 + \|S_{test}H_e\|_F^2 . \quad (35)$$

By taking the derivative of the above equation w.r.t. $S_{test}$ and zeroing the derivative, we can easily obtain $S_{test}$ as

$$S_{test} = \left(D^{*T}_{new}D^*_{new}\right)^{-1}\left(D^{*T}_{new}X_{test}H_eH_e^T\right)\left(H_eH_e^T\right)^{-1} . \quad (36)$$

After $s_{test}$ of each $x_{test}$ is obtained, the soft label information can be inferred as $W^*s_{test}$. Finally, the hard label of $x_{test}$ can be assigned as $\arg\max_{i \leq c}\left(W^*s_{test}\right)_i$, i.e., the largest entry of the soft label vector $W^*s_{test}$ determines the class assignment of $x_{test}$.

**Classification scheme two by embedding (RFDDL-e):** In this scheme, we present an efficient embedding based method. More specifically, after the coding coefficients $S$ and graph Laplacian matrix $L$ of RFDDL are obtained from training data, we learn a code-extraction projection $G$ separately by

$$G = \arg\min_G \|LGX - S\|_F^2 + \|GX\|_F^2 . \quad (37)$$

Due to the minimization $\|LGX - S\|_F^2$, learnt $G$ can potentially extract the approximate coefficients from outside new data by embedding data directly onto it. By zeroing the derivative w.r.t. $G$, we can obtain the code-extraction projection $G$ as

$$G = \left(L^TL + I\right)^{-1}L^TSX^T\left(XX^T\right)^{-1} . \quad (38)$$

After $G$ is obtained, the coefficient vector $s_{test}$ of each $x_{test}$ is denoted as $Gx_{test}$, then the soft label vector is similarly obtained as $W^*Gx_{test}$ and the hard label is assigned as $\arg\max_{i \leq c}\left(W^*Gx_{test}\right)_i$.

## E. Convergence Analysis

The variables in RFDDL are solved alternately, so we want to present its convergence analysis. Specifically, based on [41] we summarize the convergence of our RFDDL in Theorem 1.

**Theorem 1.** The optimization procedures of our RFDDL in Algorithm 1 decreases the objective function in Eq.(11) in each iteration until it converges.

*Proof:* The proposed RFDDL approach shown in Algorithm 1 can be regarded as a two-stage optimization method:

(1) **Updating $D$ and $S$ stage:** By fixing the sparse errors $E$ and $E_D$, both $D$ and $S$ can be updated by solving

$$\min_{D,S} \|(X-E)H_e - (D-E_D)LSH_e\|_F^2 + \beta\left(\|QH_e - LSH_e\|_F^2 + \|SH_e\|_F^2\right) . \quad (39)$$

Since the discriminative Laplacian matrix $L$ depends on the recovered clean dictionary $D_{new}=D-E_D$ completely, the updated $D$ and $S$ decrease the objective values of Eq.(20).

(2) **Updating $E$ and $E_D$ stage:** By fixing $D$ and $S$, both $E$ and $E_D$ can be updated by solving the following sub-problem:

$$\min_{E,E_D} \|(X-E)H_e - (D-E_D)LSH_e\|_F^2 + \alpha\left(\|E^T\|_{2,1} + \|E_D^T\|_{2,1}\right) . \quad (40)$$

With similar argument, the updated $E$ and $E_D$ decrease the objective values of Eq.(20). Because there are several blocks in RFDDL and the objective function is also non-smooth, it is not easy to prove the global convergence in theory [2], [21]. But fortunately, both above stages decrease the objective function value of Eq.(20), thus RFDDL can also decrease the original objective function value of Eq.(10) and is ensured to converge.

---

**Algorithm 1**: Robust Flexible Discriminative DL (RFDDL)

**Input:** Labeled training data $X$, discriminative sparse code matrix $Q$, dictionary size $K$ and parameters $\alpha$, $\beta$, $\lambda$.
**Initialization:** Initialize $V_0$, $\Lambda_0$ and $\mathbb{Q}_0$ as identity matrices; initialize $E_0$ and $(E_D)_0$ to be zero matrices; initialize $S_0$ as the random matrix; initialize the graph Laplacian matrix $L_0$ using Eqs.(12-13) based on the original training data; $t=0$.
*While not converge do*
1. Compute the recovered clean data as $(X_{new})_{t+1} = X - E_t$ and recovered clean dictionary as $(D_{new})_{t+1} = D_t - (E_D)_t$;
2. Update the graph Laplacian matrix $L_{t+1}$ by Eqs.(12-13);
3. Fix others, update the dictionary $D_{t+1}$ by Eq.(22);
4. Fix others, update the coefficients $S_{t+1}$ by Eq.(23);
5. Fix others, update the error $(E_D)_{t+1}$ by Eq.(26) and;
6. Fix others, update the error $E_{t+1}$ by Eq.(30);
7. Update the linear label predictor $W_{t+1}$ by Eq.(33);
8. Update the diagonal matrices $\Lambda_{t+1}$, $V_{t+1}$ and $\mathbb{Q}_{t+1}$ by Eqs.(27) (31) and (34), respectively;
9. If *converged*, stop; else, $t=t+1$ and go to the step 1.
*end while*
**Output:** Clean dictionary $D^*_{new} = (D_{new})_{t+1}$, coefficients matrix $S^* = S_{t+1}$, Laplacian matrix $L^* = L_{t+1}$ and classifier $W^* = W_{t+1}$.

---

## IV. DISCUSSION: RELATIONSHIP ANALYSIS

We mainly discuss the connections and differences between our RFDDL and other related work in this part.

### A. Relation to D-KSVD and LC-KSVD

Recall the objective function of RFDDL in Eq.(11), if the ideal conditions that the original data and computed discriminative dictionary are absolutely clean without any noise and errors are satisfied, i.e., $E=0$ and $E_D=0$, and suppose that the bias $b_1, b_2, b_3$ are zeros, we can easily have the reduced problem:

$$\min_{\substack{D,S,L,b_1,\\b_2,b_3,W}} \|X + b_1e^T - DLS\|_F^2 + \beta\left(\|Q + b_2e^T - LS\|_F^2 + \|SH_e\|_F^2\right) \\ + \gamma\left(\|H + b_3e^T - WLS\|_F^2 + \|W^T\|_{2,1}\right) . \quad (41)$$

By comparing the above problem with that of LC-KSVD in Eq.(2), we can find that one difference is that RFDDL embeds sparse codes into a discriminative Laplacian matrix $L$ to form locality-constrained structure preserving coefficients, while the LC-KSVD uses a transformation $A$ that is not clearly defined to enforce transformed codes $AS$ to approximate $Q$. The second difference is that RFDDL minimizes the reconstruction error, discriminative sparse code error and classification error in a flexible manner, while LC-KSVD impose the hard constraints on them, which may produce inaccurate representations. The third one is that LC-KSVD clearly uses the constraint $\|s_i\|_0 \leq T_0$ to ensure the sparsity of learnt coefficients, while our RFDDL regularizes the Frobenius-norm on the centered coefficients for the efficient coding. The fourth one is that our RFDDL clearly associates the Laplacian matrix $L$ with the coding coefficients, dictionary and classifier jointly for enhancing the inter-class discrimination. It is worth noting that the problems of RFDDL and LC-KSVD are also equivalent to some extent when the following conditions are satisfied: (1) $A$ is fixed as $L$ in each iteration; (2) the bias $b_1, b_2, b_3$ are zeros, i.e., the reconstructions are accurate; (3) the factor $T_0$ in $\|s_i\|_0 \leq T_0$ is set to a large value and in such case the learnt $s_i$ of each sample lose the sparse properties; (4) the graph Laplacian $L$ associated with the data reconstruction and classification errors are removed. Based on these analyses, if we further remove the discriminative sparse code error term, D-KSVD is also equivalent to our RFDDL. Hence, by removing the useful constraints and regularization, both D-KSVD and LC-KSVD will be inferior to RFDDL for classification in theory, which will be verified by simulations.

*B. Relation to LCLE-DL*

We also describe the connections between LCLE-DL and our RFDDL. Recalling the objective function of LCLE-DL, if we set $\gamma$ associated with the term $\|S-V\|_2^2$ to $+\infty$, i.e., $\|S-V\|_2^2 \to 0$ or $S=V$, the objective function of LCLE-DL can be reduced to

$$\min_{D,S,L} \|X - DS\|_2^2 + \alpha tr(S^T LS) + \beta tr(S^T US)$$
$$s.t. \|d_i\|^2 = 1, i = 1,\dots,K \tag{42}$$

which means that the label constraint and the locality constraint are all regularized on the coefficient matrix $S$. To facilitate the comparison, by setting the bias $b_1, b_3$ to be zeros and $\gamma = 0$ (i.e., the classifier $W$ is not jointly learnt) in the formulation of our RFDDL in Eq.(11), we can simplify the problem as

$$\min_{D,S,L,E,E_D,b_2} \|(X-E)-(D-E_D)LS\|_F^2 + \alpha\left(\|E^T\|_{2,1} + \|E_D^T\|_{2,1}\right)$$
$$+ \beta\left(\|Q+b_2 e^T - LS\|_F^2 + \|SH_e\|_F^2\right). \tag{43}$$

By comparing the above two problems, we can find that: LCLE-DL uses $tr(S^T LS)$ to preserve the locality of the learned dictionary and uses the label embedding term $tr(S^T US)$ instead of classification error to encourage intra-class atoms to have the similar profiles, while RFDDL uses the discriminative flexible sparse code error term $\|Q+b_2 e^T - LS\|_F^2$ to preserve the locality of learned dictionary and encourage the subspace structures or label consistency of learned atoms and coding coefficients to be preserved at the same time. That is, the purposes of them are similar, but they employs different strategies. As a result, the above formulation can be regarded as a robust enhanced variant of LCLE-DL. Moreover, setting $\gamma = 0$ means that the codes of LCLE-DL cannot be ensured to be optimal for classification, although LCLE-DL uses the label embedding of atoms to force the coefficient matrix to be block-diagonal. Hence, RFDDL can also potentially outperform LCLE-DL for data classification by incorporating the subspace recovery and enhanced locality.

## V. SIMULATION RESULTS AND ANALYSIS

In this section, we mainly evaluate RFDDL for representation and classification of images and documents. The performance of RFDDL is compared with those of 14 related algorithms, that is, SRC [20], KSVD [1], D-KSVD [25], FDDL [22], DPL [5], LC-KSVD1 [9], LC-KSVD2 [9], JDDRDL [23], SDR [24], SVGDL [34], DLSI [35], ADDL [32], COPAR [42], LCLE-DL [33], and LRSDL [43]. Nine public databases, including four face databases: ORL (*http://www.uk.research.att.com/facedatabase.html*), UMIST [6], MIT CBCL [37] and CMU PIE [48], three object image databases: Caltech101 [14], ETH80 [11] and COIL20 [49], and two text databases: TDT2 (*https://www.nist.gov/speech/tests/tdt/tdt98/index.htm*) and RCV1 [50]. Detailed information of used datasets are described in Table I and some image examples are illustrated in Fig.1. The parameters of each evaluated algorithm are all carefully chosen for fair comparison. As is common practice, all images of ORL, MIT CBCL, CMU PIE, ETH80 and COIL20 are resized into 32×32 pixels due to efficiency. Thus, each image corresponds to a data point in a 1024-dimensional space. For the inductive classification, we split each database into a training set and a test set, where the training set is used for DL and coefficients coding, and the test set is to evaluate the classification accuracy. KSVD uses the same classification approach as SRC [20]. The Gaussian kernel width is set by the estimation approach of [44] in LCLE-DL and the nearest neighbor number is set to 7 [45] for LCLE-DL and RFDDL. The dimension of JDDRDL and SDR is reduced to $d=c-1$ [45], [46]. We perform all simulations on a PC with Intel Core i7-6700 CPU @ 3.4 GHz 3.4GHz 8G.

TABLE I. DESCRIPTIONS OF EVALUATED DATABASES.

| Dataset Name | # Samples | # Dim | # Classes |
|---|---|---|---|
| MIT CBCL (face) | 3240 | 1024 | 10 |
| UMIST (face) | 1012 | 1024 | 20 |
| CMU PIE (face) | 11554 | 1024 | 68 |
| Caltech101 (object) | 9144 | 3000 | 102 |
| ETH80 (object) | 3280 | 1024 | 80 |
| COIL20 (object) | 1440 | 1024 | 20 |
| TDT2 (text) | 9394 | 36771 | 30 |
| RCV1 (text) | 9625 | 29992 | 4 |

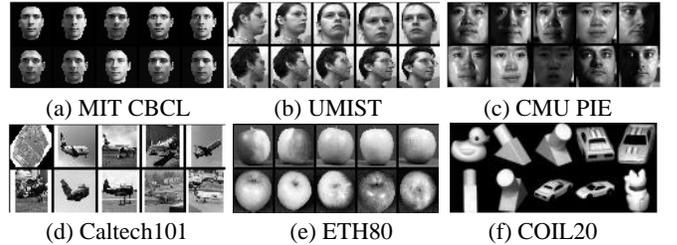

(a) MIT CBCL   (b) UMIST   (c) CMU PIE
(d) Caltech101   (e) ETH80   (f) COIL20
**Fig. 1:** Some image examples of evaluated image databases.

*A. Visual Image Analysis by Visualization*

We present some results to visualize the recovered clean data and clean dictionary in the real cases. Two databases, i.e., CMU PIE and ORL, are evaluated. CMU PIE database has 68 persons with 41368 images under varying pose, illumination and facial expression. Following [1], 170 near frontal images per person are used. We choose the five near frontal poses (C05, C07, C09, C27, and C29) and use all images under different illuminations and facial expressions [44]. We randomly choose 5 face images per class to form the data matrix $X$. The number of dictionary atoms is set to 340 and 50 for CMU PIE and ORL databases, respectively. To observe the denoising effect, random Gaussian noise is added manually into the data matrix $X$ by using

$$X' = X + \sqrt{variance} \times randn(size(X)). \tag{44}$$

We first visualize the effects of recovering the original face images, and mainly show several images as examples for clear observation in Fig.2, where we show the original noisy images $X'$, recovered images $X_{new}$ and error images $E$. We also quantify the denoising results by computing the signal-to-noise ratio (SNR) and root-mean-square error (RMSE) based on the noisy and denoised images, respectively. In this study, the variance is set to 500. We can observe that the recovered images $X_{new}$ has less noise than noisy images $X'$, i.e., the recovery can remove the underlying noise from original images effectively. Note that the coefficients obtained by performing DL in recovered space will be more accurate than in original image space empirically. The quantitative evaluations also demonstrate the effectiveness of the denoising process by subspace recovery, because the recovered images $X_{new}$ obtain higher SNR values and smaller RMSE values than those from the original noisy images.

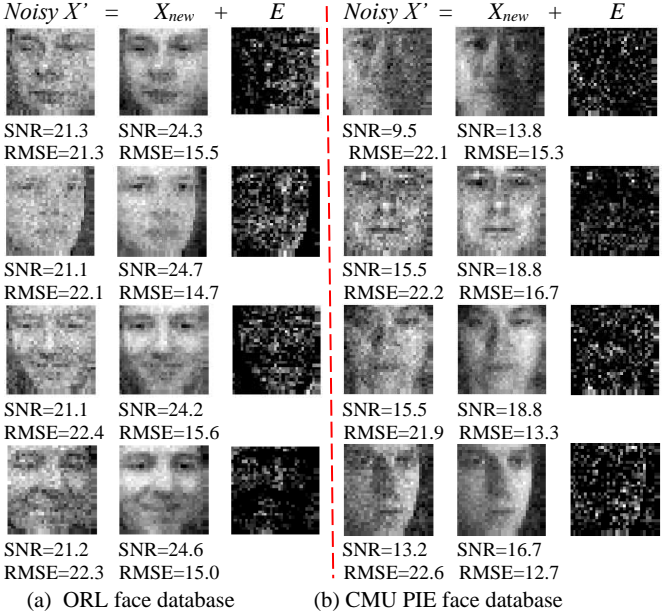

**Fig. 2:** Visualization of the recovered images on two face databases.

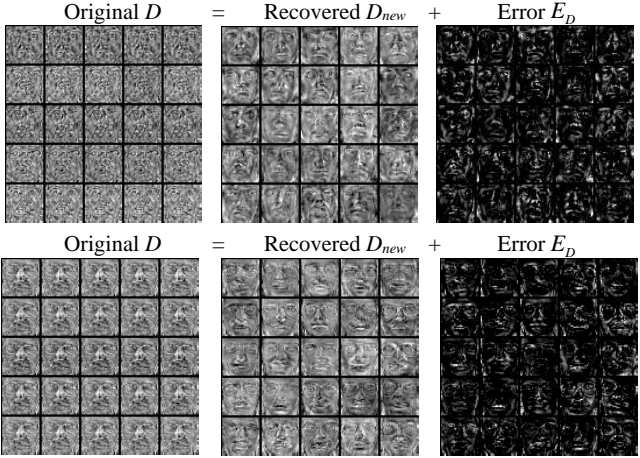

**Fig.3:** Visualization of the recovered dictionary on CMU PIE database (first row) and ORL face database (second row).

Then, we visualize the recovered dictionary in Fig.3, where we show 25 images for clear observation. We can find that the recovered dictionary $D_{new}$ has less noise than original dictionary $D$. More importantly, the learnt $D$ is vague and is difficult to be distinguished for inter-class images due to poor discriminating power. On the contrary, the recovered $D_{new}$ can capture more face details than $D$, which is benefit for learning discriminative coefficients, i.e., the dictionary recovery can also remove the underlying noise. Note that the coefficients by clean dictionary $D_{new}$ will be intuitively more accurate than original dictionary.

### B. Convergence Results

We present some convergence results of RFDDL by illustrating the objective function values. Six databases, i.e., MIT CBCL, UMIST, ETH80, Caltech101, TDT2 and RCV1, are applied for evaluations. A subset of the original TDT2 corpus is employed, where those documents appearing in two or more categories were removed, and the largest 30 categories were kept, leaving us with 9,394 documents in total. RCV1 text database contains information of topics, regions and industries for each document and a hierarchical structure for topics and industries. A set of 9,625 documents with 29,992 distinct words is applied in this study, including the categories "C15," "ECAT," "GCAT," and "MCAT," with each having 2,022, 2,064, 2,901, and 2,638 documents. For each database, we choose 5 images from each class for training and DL. The convergence results are shown in Fig.4, from which we can find that the objective function value of our RFDDL in the iterations is non-increasing and usually converges to a constant. RFDDL also converge fast.

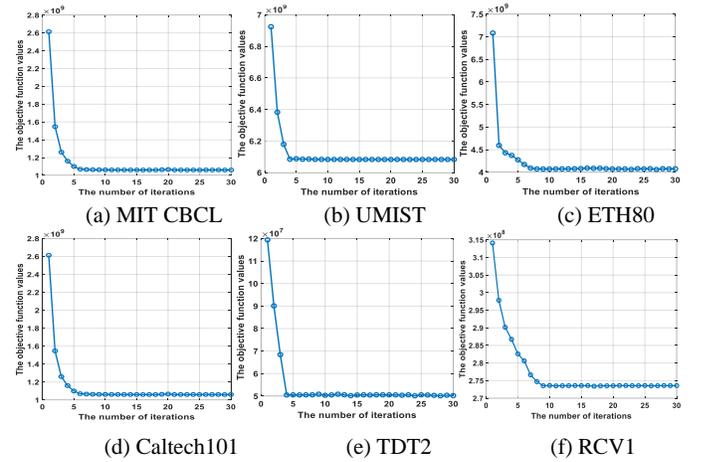

**Fig. 4:** The convergence results of our RFDDL on some databases.

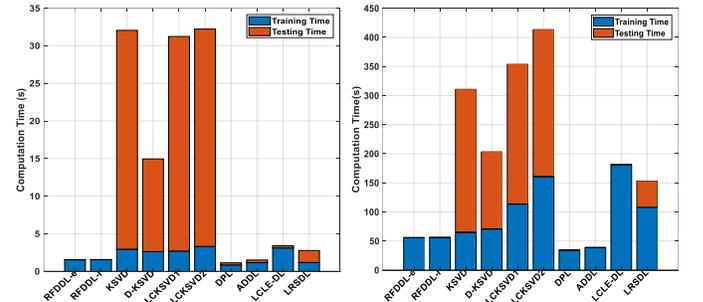

**Fig. 5:** Quantitative comparison of computational time of each method on MIT CBCL (left) and CMU PIE (right).

### C. Comparison of Computational Time

We evaluate the actual computational time, including both the training and testing time of each algorithm. Two face datasets, i.e., MIT CBCL [37] and CMU PIE [48], are employed for the evaluations. For each database, we select 20 samples from each class randomly as training set and test on the rest. The averaged computational time of each approach over 20 different runs is shown in Fig. 5. From the running time, we can find that: (1) the overall computational time of our RFDDL-r and RFDDL-e are comparable to those of DPL and ADDL, since RFDDL uses the efficient Frobenius norm instead of the costly $L_0/L_1$-norm for coefficient coding in training stage; (2) KSVD, D-KSVD and LC-KSVD spend far more test time and overall computational time than other models, since they have to involve a separable costly sparse reconstruction process for each new data to obtain its coefficient vector and then perform classification over it. On the contrary, the costly sparse reconstruction process is avoided in DPL, ADDL and our RFDDL, and specifically they used the efficient inclusion schemes to handle new test data.

### D. Application to Face Recognition

We evaluate each model for face representation and recognition on three public real databases, i.e., MIT CBCL, UMIST and CMU PIE. The recognition result of each method is averaged over 10 random splits of training and test images. The sparsity constraint factor is set to 30 for each $L_0$-norm based method.

*1) Results on MIT CBCL.* In this study, we randomly select 4 images from each person for training, while the rest are used for testing. We set $\alpha=10^2$, $\beta=10^6$, $\lambda=10^8$ for our RFDDL-e and RFDDL-r. The number of atoms is set as its maximum for each method if without special remarks. Table II describes the face recognition result and the highest two records are highlighted in bold, from which we can find that our RFDDL-r and RFDDL-e obtain higher accuracies than other methods. The improvement by our models can be attributed to encoding the flexible sparse reconstruction error, discriminative flexible sparse code error and flexible classification error more accurately and recovering underlying clean data and atom subspaces for representations jointly. LC-KSVD2, LCLE-DL, LRSDL and ADDL can also perform better than other methods. The confusion matrices for RFDDL-r and RFDDL-e are shown in Figs.6 (a) and (b).

TABLE II.
AVERAGED FACE RECOGNITION RATES ON MIT CBCL.

| Evaluated Methods | Accuracy (%) |
|---|---|
| SRC(4 items per person) | 93.1 |
| KSVD(4 items per person) | 93.2 |
| D-KSVD(4 items per person) | 95.8 |
| JDDRDL(4 items per person) | 96.0 |
| SDR(4 items per person) | 94.3 |
| FDDL(4 items per person) | 96.0 |
| LC-KSVD1(4 items per person) | 96.5 |
| LC-KSVD2(4 items per person) | 97.3 |
| DLSI(4 items per person) | 94.1 |
| SVGDL(4 items per person) | 94.6 |
| DPL(4 items per person) | 96.3 |
| ADDL(4 items per person) | 97.7 |
| COPAR(4 items per person) | 97.1 |
| LCLE-DL(4 items per person) | 97.4 |
| LRSDL(4 items per person) | 97.6 |
| **RFDDL-r(4 items per person)** | **98.5** |
| **RFDDL-e(4 items per person)** | **98.7** |

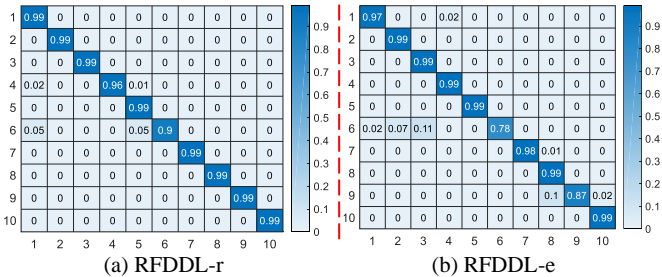

(a) RFDDL-r    (b) RFDDL-e

**Fig. 6:** Confusion matrices of our proposed methods on MIT CBCL.

*2) Results on UMIST.* For this study, we randomly select 5 images from each person as the training set and test on the other images. We set $\alpha=10^2$, $\beta=10^4$, $\lambda=10^8$ for RFDDL-e and RFDDL-r. The recognition results of each algorithm are described in Table III. We find from the results that our RFDDL-r and RFDDL-e methods can obtain the enhanced recognition results than other compared algorithms. The face recognition results of ADDL, COPAR and LRDSL are comparable with each other, and are superior to the other remaining algorithms. DLSI and JDDRDL obtain the worse results on this face database.

TABLE III.
AVERAGED FACE RECOGNITION RATES ON UMIST.

| Evaluated Methods | Accuracy (%) |
|---|---|
| SRC(5 items per person) | 87.4 |
| KSVD(5 items per person) | 87.7 |
| D-KSVD(5 items per person) | 87.2 |
| JDDRDL(5 items per person) | 85.3 |
| SDR(5 items per person) | 88.1 |
| FDDL(5 items per person) | 87.2 |
| LC-KSVD1(5 items per person) | 87.8 |
| LC-KSVD2(5 items per person) | 88.6 |
| DLSI(5 items per person) | 84.9 |
| SVGDL(5 items per person) | 87.7 |
| DPL(5 items per person) | 88.9 |
| ADDL(5 items per person) | 89.3 |
| COPAR(5 items per person) | 89.2 |
| LCLE-DL(5 items per person) | 88.5 |
| LRSDL(5 items per person) | 89.8 |
| **RFDDL-r(5 items per person)** | **90.7** |
| **RFDDL-e(5 items per person)** | **91.4** |

TABLE IV.
AVERAGED RECOGNITION RATES ON CMU PIE UNDER DIFFERENT NUMBERS OF TRAINING SAMPLES.

| Evaluated Methods | 5 train/class (5 items) Acc. (%) | 10 train/class (8 items) Acc. (%) | 15 train/class (12 items) Acc. (%) | 20 train/class (16 items) Acc. (%) |
|---|---|---|---|---|
| SRC | 53.5 | 68.9 | 73.4 | 78.5 |
| KSVD | 51.5 | 56.1 | 69.4 | 75.6 |
| D-KSVD | 53.8 | 59.7 | 67.2 | 77.8 |
| JDDRDL | 48.3 | 66.8 | 72.4 | 75.3 |
| SDR | 43.1 | 65.4 | 70.2 | 73.6 |
| FDDL | 40.3 | 58.8 | 66.1 | 75.3 |
| LC-KSVD1 | 52.4 | 60.5 | 68.5 | 77.3 |
| LC-KSVD2 | 54.2 | 71.4 | 77.5 | 79.5 |
| DLSI | 39.4 | 56.3 | 60.8 | 67.5 |
| SVGDL | 54.2 | 70.6 | 74.3 | 78.2 |
| DPL | 52.3 | 60.1 | 68.7 | 74.3 |
| ADDL | 54.5 | 78.3 | 85.2 | 89.2 |
| COPAR | 54.2 | 71.0 | 81.0 | 86.8 |
| LCLE-DL | 54.3 | 71.5 | 77.8 | 80.1 |
| LRSDL | 53.3 | 72.0 | 81.7 | 87.2 |
| **RFDDL-r** | **55.2** | **79.8** | **86.1** | **89.3** |
| **RFDDL-e** | **56.5** | **81.5** | **88.3** | **90.2** |

*3) Results on CMU PIE.* In this section, we used the principal component features in [32] for face recognition. We evaluate recognition result of each algorithm under different numbers of training samples by randomly selecting 5, 10, 15 and 20 images from each person as the training set and test on the rest images. We also use different numbers of dictionary items for different training samples evaluate the results, which corresponds to an average of 5, 8, 12 and 16 items per person. The parameters $\alpha=10^2$, $\beta=10^4$, $\lambda=10^2$ are set for our RFDDL-e and $\alpha=10^2$, $\beta=10^8$ $\lambda=10^8$ for RFDDL-r. The averaged results are shown in Table IV. We find that the results of each algorithm can be improved as the number of training data increases. RFDDL-e performs the best among all compared methods, followed by RFDDL-r and ADDL. LRSDL and COPAR can also work well.

*E. Application to Object Recognition*

We evaluate RFDDL-r and RFDDL-e for object representation and recognition on three public real databases, i.e., Caltech101, COIL20 and ETH80. The sparsity constraint factor is fixed to 30 for Caltech101 and 10 for COIL20 and ETH80.

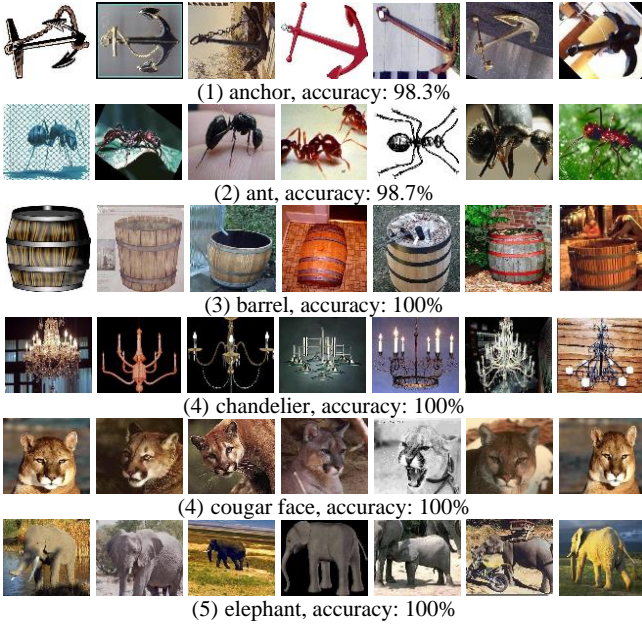

(1) anchor, accuracy: 98.3%
(2) ant, accuracy: 98.7%
(3) barrel, accuracy: 100%
(4) chandelier, accuracy: 100%
(4) cougar face, accuracy: 100%
(5) elephant, accuracy: 100%

**Fig. 7:** Example images from classes with high recognition accuracy from the Caltech101 object database.

TABLE V.
AVERAGED OBJECT RECOGNITION RATES ON CALTECH101.

| Evaluated Methods | Accuracy (%) |
|---|---|
| SRC(30 items per class) | 70.7 |
| KSVD(30 items per class) | 73.2 |
| D-KSVD(30 items per class) | 73.0 |
| JDDRDL(30 items per class) | 67.8 |
| SDR(30 items per class) | 70.2 |
| FDDL(30 items per class) | 73.1 |
| LC-KSVD1(30 items per class) | 73.4 |
| LC-KSVD2(30 items per class) | 73.6 |
| DLSI(30 items per class) | 70.1 |
| SVGDL(30 items per class) | 72.6 |
| DPL(30 items per class) | 73.9 |
| ADDL(30 items per class) | 74.2 |
| COPAR(30 items per class) | 74.0 |
| LCLE-DL(30 items per class) | 73.5 |
| LRSDL(30 items per class) | 74.2 |
| RFDDL-r(30 items per class) | **74.3** |
| RFDDL-e(30 items per class) | **74.6** |

*1) Results on Caltech101.* In this experiment, we apply the spatial pyramid features of [9] and randomly select 30 images from each category for training and test on the rest. We set parameters $\alpha=10^2, \beta=10^4, \lambda=10^8$ for RFDDL-e and RFDDL-r. The averaged recognition results are described in Table V. We can find that our RFDDL-e and RFDDL-r can deliver better accuracies than other models. ADDL and LRSDL also perform well by delivering competitive results to our criteria. JDDRDL is the worst one. The performance of DPL and COPAR is better than KSVD, FDDL, LCLE-DL, LC-KSVD1 and LC-KSVD2. In addition, Fig.7 illustrates the example images from the object classes with high recognition accuracies.

*2) Results on COIL20.* In this study, we randomly select 10 images from each class for training and use the rest for testing. We set $\alpha=10^2, \beta=10^4, \lambda=10^8$ for RFDDL-e and RFDDL-r. From the recognition result in Table VI, we find that our RFDDL-e obtains the best record, and our RFDDL-r delivers the highly comparable results to ADDL, LRSDL and COPAR.

TABLE VI.
AVERAGED OBJECT RECOGNITION RATES ON COIL20.

| Evaluated Methods | Accuracy (%) |
|---|---|
| SRC(10 items per class) | 84.0 |
| KSVD(10 items per class) | 84.1 |
| D-KSVD(10 items per class) | 84.6 |
| JDDRDL(10 items per class) | 83.2 |
| SDR(10 items per class) | 83.4 |
| FDDL(10 items per class) | 83.6 |
| LC-KSVD1(10 items per class) | 83.2 |
| LC-KSVD2(10 items per class) | 85.5 |
| DLSI(10 items per class) | 85.8 |
| SVGDL(10 items per class) | 84.3 |
| DPL(10 items per class) | 84.2 |
| ADDL(10 items per class ) | 85.8 |
| COPAR(10 items per class) | 85.8 |
| LCLE-DL(10 items per class) | 85.2 |
| LRSDL(10 items per class) | 86.2 |
| RFDDL-r(10 items per class) | **87.3** |
| RFDDL-e(10 items per class) | **87.8** |

TABLE VII.
AVERAGED OBJECT RECOGNITION RATES ON ETH80.

| Evaluated Methods | Accuracy (%) |
|---|---|
| SRC(10 items per class) | 88.3 |
| KSVD(10 items per class) | 87.9 |
| D-KSVD(10 items per class) | 89.2 |
| JDDRDL(10 items per class) | 85.8 |
| SDR(10 items per class) | 85.1 |
| FDDL(10 items per class) | 87.4 |
| LC-KSVD1(10 items per class) | 87.9 |
| LC-KSVD2(10 items per class) | 88.6 |
| DLSI(10 items per class) | 86.9 |
| SVGDL(10 items per class) | 88.3 |
| DPL(110 items per class) | 89.7 |
| ADDL(10 items per class) | 89.9 |
| COPAR(10 items per class) | 89.4 |
| LCLE-DL(10 items per class) | 88.7 |
| LRSDL(10 items per class) | 89.6 |
| RFDDL-r(10 items per class) | **90.3** |
| RFDDL-e (10 items per class) | **91.2** |

*3) Results on ETH80.* In this study, we mainly consider an eight-class object categorization problem, i.e., each of the eight big categories is treated as a single class. We use the features of [32], randomly select 10 images per category for training and test on the rest. $\alpha=10^2, \beta=10^6, \lambda=10^8$ are set for RFDDL-e and $\alpha=10^4, \beta=10^4, \lambda=10^4$ are set for our RFDDL-r. We describe the averaged results in Table VII, from which we can find that our algorithms outperform other compared methods. The confusion matrices for RFDDL-r and RFDDL-e are shown in Fig.8, from which we find that most of the confusion occurs between cow, dog, horse and tomato. We also evaluate the recognition rates of RFDDL-e for individual classes in Fig. 9, from which we can obtain the similar conclusion about the performance as Fig.8.

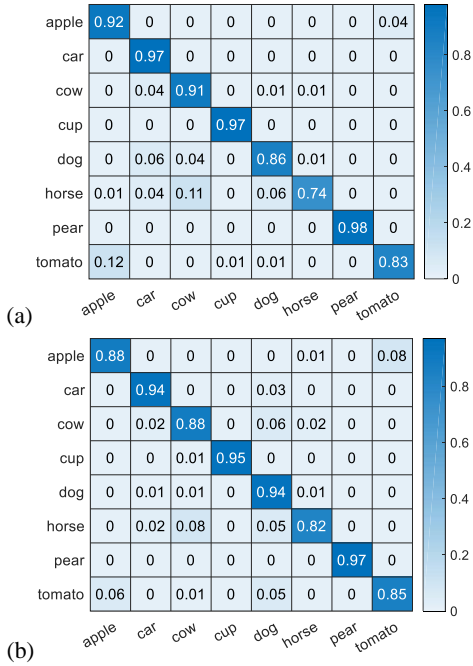

**Fig. 8:** Confusion matrices of RFDDL-r (a) and RFDDL-e (b) on the ETH80 database.

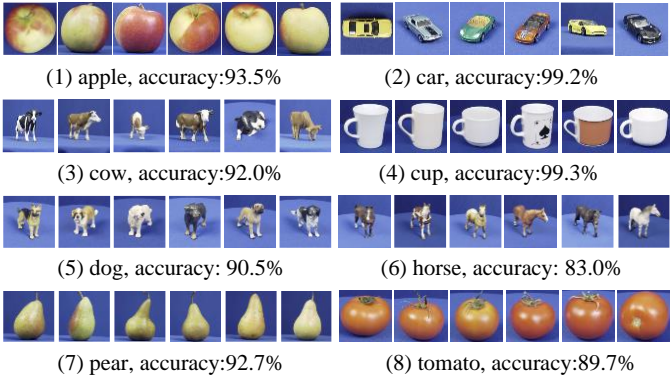

(1) apple, accuracy:93.5%  (2) car, accuracy:99.2%
(3) cow, accuracy:92.0%  (4) cup, accuracy:99.3%
(5) dog, accuracy: 90.5%  (6) horse, accuracy: 83.0%
(7) pear, accuracy:92.7%  (8) tomato, accuracy:89.7%

**Fig. 9:** Example images with accuracy from the ETH80 database.

### F. Application to Text Categorization

We also evaluate each algorithm for categorizing texts based on two popular document databases, i.e., TDT2 and RCV1.

*1) Results on TDT2.* For the consideration of computational efficiency, we use the Principal Component Analysis (PCA)[47] as a preprocessing step to reduce the number of dimension to 3000. We randomly select 10 text data per class for training and test on the rest. We set $\alpha=10^2$, $\beta=10^2$ and $\lambda=10^8$ for RFDDL-r. $\alpha=10^6$, $\beta=10^2$ and $\lambda=1$ are used for RFDDL-e. From the text categorization results in Table VIII, we can find that RFDDL-r and RFDDL-e obtain the comparable categorization accuracies, and both are superior to other remaining methods.

*2) Results on RCV1.* In this experiment, we also extract the principal components features with the dimension being 2000 by PCA. $\alpha=10^2, \beta=10^4, \lambda=10^8$ are set for RFDDL-e and $\alpha=10$, $\beta=0.01, \lambda=10^4$ for RFDDL-r. In this study, we train on 10, 20, 40 and 80 samples per category with different dictionary items per class and test on the rest. The final rates are reported as the average of each run in Table IX. We find that: (1) the increasing numbers of training samples per class can clearly enhance the performance; (2) RFDDL-e and RFDDL-r can deliver better results than other methods. ADDL and LRSDL also perform well by delivering highly competitive results to our methods.

TABLE VIII.
AVERAGED TEXT CATEGORIZATION RESULTS ON TDT2.

| Evaluated Methods | Accuracy (%) |
|---|---|
| SRC(10 items per class) | 75.1 |
| KSVD(10 items per class) | 68.0 |
| D-KSVD(10 items per class) | 73.6 |
| JDDRDL(10 items per class) | 76.4 |
| SDR(10 items per class) | 79.1 |
| FDDL(10 items per class) | 69.3 |
| LC-KSVD1(10 items per class) | 77.8 |
| LC-KSVD2(10 items per class) | 79.9 |
| DLSI(10 items per class) | 70.2 |
| SVGDL(10 items per class) | 80.2 |
| DPL(10 items per class) | 77.3 |
| ADDL(10 items per class) | 77.8 |
| COPAR(10 items per class) | 75.5 |
| LCLE-DL(10 items per class) | 78.6 |
| LRSDL(10 items per class) | 76.6 |
| RFDDL-r(10 items per class) | **84.0** |
| RFDDL-e(10 items per class) | **83.4** |

TABLE IX.
AVERAGED TEXT CATEGORIZATION RATES ON RCV1 UNDER DIFFERENT NUMBERS OF TRAINING SAMPLES.

| Evaluated Methods | 10 train (8 items) Acc. (%) | 20 train (16 items) Acc. (%) | 40 train (30 items) Acc. (%) | 80 train (60 items) Acc. (%) |
|---|---|---|---|---|
| SRC | 67.4 | 72.1 | 77.9 | 82.6 |
| KSVD | 68.9 | 74.5 | 78.4 | 81.5 |
| D-KSVD | 71.3 | 79.3 | 83.5 | 87.9 |
| JDDRDL | 69.2 | 76.3 | 83.1 | 88.1 |
| SDR | 60.5 | 70.1 | 73.5 | 76.8 |
| FDDL | 56.2 | 68.9 | 77.4 | 81.2 |
| LC-KSVD1 | 72.8 | 80.6 | 84.1 | 88.4 |
| LC-KSVD2 | 72.9 | 80.7 | 85.4 | 88.9 |
| DLSI | 69.8 | 76.3 | 85.2 | 88.5 |
| SVGDL | 72.5 | 79.4 | 86.7 | 89.2 |
| DPL | 69.5 | 79.3 | 82.1 | 85.7 |
| ADDL | 77.0 | 81.6 | 87.5 | 89.8 |
| COPAR | 74.8 | 80.8 | 86.4 | 89.4 |
| LCLE-DL | 71.8 | 80.5 | 83.5 | 86.4 |
| LRSDL | 76.2 | 82.1 | 87.8 | 90.2 |
| RFDDL-r | **77.4** | **82.8** | **88.5** | **90.4** |
| RFDDL-e | **80.5** | **85.4** | **88.9** | **92.0** |

TABLE X.
INFORMATION ON TRAINING NUMBER AND DICTIONARY SIZE.

| Database | #Train per class | Dictionary size |
|---|---|---|
| MIT CBCL | 4 | 1c, 2c, 3c, 4c |
| UMIST | 5 | 1c, 2c, ...,5c |
| Caltech101 | 30 | 6c, 12c, ..., 30c |
| ETH80 | 10 | 2c, 4c,..., 10c |
| TDT2 | 10 | 2c, 4c, ..., 10c |
| RCV1 | 20 | 4c, 8c,..., 20c |

### G. Classification against Varying Dictionary Sizes

We also investigate the effects of various dictionary sizes on the classification results. In this experiment, two face databases (UMIST and MIT CBCL), two object databases (ETH80 and Caltech101) and two text datasets (TDT2 and RCV1) are used for the evaluations. For each database, the number of training samples per class and used dictionary sizes are shown in Table X, where *c* is the total number of classes. The classification results under various dictionary sizes are shown in Fig.10. We can observe that: (1) The performance of each algorithm can be improved with the increasing dictionary sizes; (2) RFDDL-e and RFDDL-r can generally perform better than other methods across all dictionary sizes in most cases. RFDDL-e obtains the highest accuracies than RFDDL-r and other methods in most cases, except for TDT2. On TDT2, the results of RFDDL-e and RFDDL-r are highly competitive with each other.

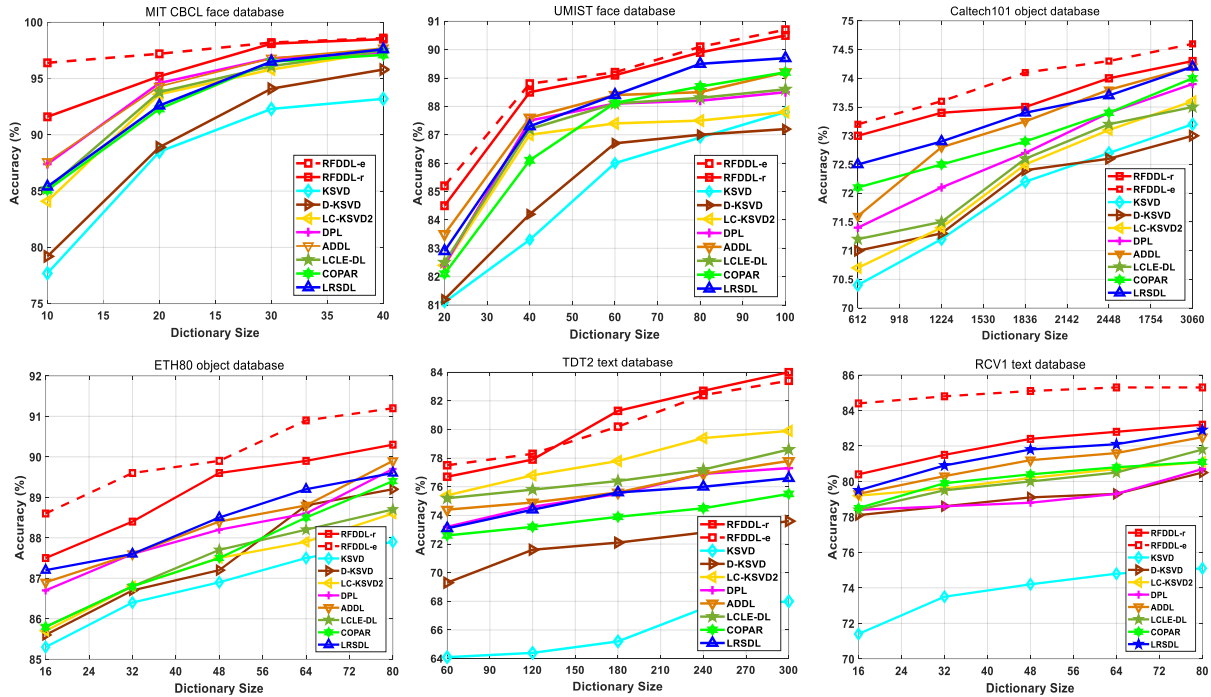

**Fig. 10:** Classification accuracy vs. varying dictionary sizes based on two face, two object and two text databases.

### H. Noisy Image Recognition Against Corruptions

We explore the robustness properties of each method against corrupted data. Two face databases (UMIST and MIT CBCL) and two object databases (ETH80 and COIL20) are applied. To corrupt image data, random Gaussian noise is included.

### (1) Face Recognition with Corruptions

In this study, the number of labeled training images per class is set to 8 and 5 for UMIST and MIT CBCL, respectively. The numbers of dictionary items are set to 6 and 4 for UMIST and MIT CBCL, respectively. The noisy face recognition results are shown in Fig.11, where the variance of random Gaussian noise is set to 50, 100, ..., 500, and some examples of noisy images are also shown. We can find that: (1) the classification result is decreased when the noise level is increased, i.e., the corruptions clearly have the negative effects on the recognition results; (2) our RFDDL-e and RFDDL-r can deliver the higher accuracies than other methods under different noise levels. Specifically, RFDDL-e and RFDDL-r degrades slower than other methods when the noise variance increases. That is, our models are less sensitive to the noise and corruptions than the other methods, which can be attributed to the integrated double recovery on data and dictionary to exploit the underlying clean subspaces, and the mechanism to associate the locality with the codes and classifier to produce accurate coefficients and label predictions. LRSDL also works well by obtaining better results, since it also considers the low-rank regularization on the dictionary.

### (2) Object Recognition with Corruptions

We also explore the noisy object recognition tasks on ETH80 and COIL20 databases to investigate the robustness property of each algorithm to noise and corruptions in the object images. We still fix the number of labeled training samples of each class and number of the dictionary items to investigate the effects of different noise levels in terms of variance on the performance. Specifically, the numbers of labeled training samples per class are set to 10 for ETH80 and COIL20, respectively. The number of dictionary items is set to 80 and 200 for ETH80 and COIL20. The numerical noisy object recognition results on ETH80 and COIL20 are shown in Tables XI and XII. We can find that: (1) the increasing noise variance clearly decreases the recognition results; (2) RFDDL-e and RFDDL-r deliver higher accuracy than other methods in most cases. Specifically, the performance improvement of RFDDL-e and RFDDL-r over other methods is more obvious when the noise level is relatively higher.

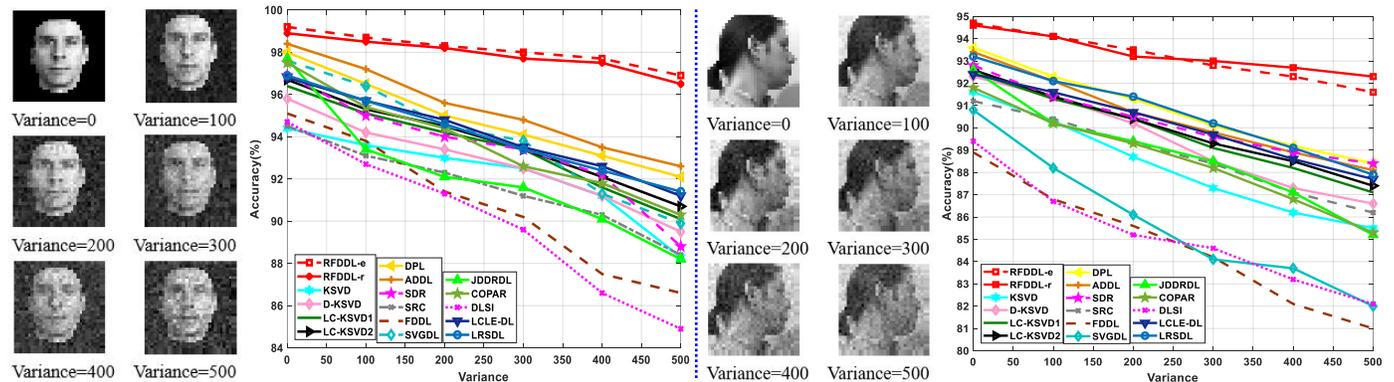

**Fig. 11:** Classification performance vs. varying variance on the MIT CBCL (left) and UMIST (right) face databases.

TABLE XI.
AVERAGED OBJECT RECOGNITION RATES ON NOISY ETH80 DATABASE WITH VARYING VARIANCE.

| Setting Method | ETH80 var=2 | ETH80 var=8 | ETH80 var=50 | ETH80 var=100 |
|---|---|---|---|---|
| SRC | 88.3 | 82.6 | 65.4 | 45.9 |
| KSVD | 86.9 | 84.7 | 66.9 | 57.9 |
| D-KSVD | 88.0 | 85.0 | 65.5 | 55.9 |
| JDDRDL | 87.9 | 85.3 | 71.4 | 59.1 |
| SDR | 86.7 | 84.7 | 69.1 | 56.8 |
| FDDL | 88.7 | 85.9 | 66.2 | 57.5 |
| LC-KSVD1 | 87.9 | 84.1 | 68.7 | 57.0 |
| LC-KSVD2 | 88.6 | 84.7 | 69.0 | 57.9 |
| DLSI | 86.6 | 84.2 | 65.1 | 50.3 |
| SVGDL | 83.7 | 82.1 | 71.1 | 62.7 |
| DPL | 88.9 | 85.9 | 70.2 | 59.8 |
| ADDL | 89.4 | 86.7 | 69.9 | 59.6 |
| COPAR | 88.3 | 84.2 | 70.5 | 60.1 |
| LCLE-DL | 88.4 | 84.2 | 68.1 | 58.2 |
| LRSDL | 88.9 | 85.9 | 69.8 | 59.4 |
| RFDDL-r | **89.5** | **86.9** | **72.8** | **66.1** |
| RFDDL-e | **91.0** | **88.4** | **75.0** | **68.1** |

TABLE XII.
AVERAGED OBJECT RECOGNITION RATES ON NOISY COIL20 DATABASE WITH VARYING VARIANCE.

| Setting Method | COIL20 var=2 | COIL20 var=8 | COIL20 var=50 | COIL20 var=100 |
|---|---|---|---|---|
| SRC | 83.5 | 83.1 | 82.6 | 81.2 |
| KSVD | 83.7 | 83.2 | 82.5 | 81.6 |
| D-KSVD | 84.1 | 83.5 | 83.1 | 82.1 |
| JDDRDL | 83.0 | 82.6 | 81.9 | 80.8 |
| SDR | 83.1 | 82.7 | 82.3 | 81.6 |
| FDDL | 83.2 | 82.5 | 82.1 | 80.5 |
| LC-KSVD1 | 82.7 | 82.3 | 81.6 | 80.2 |
| LC-KSVD2 | 85.5 | 84.2 | 83.6 | 81.4 |
| DLSI | 85.8 | 85.6 | 84.7 | 82.9 |
| SVGDL | 84.3 | 83.9 | 83.0 | 81.7 |
| DPL | 84.2 | 83.8 | 83.1 | 81.2 |
| ADDL | 85.8 | 85.5 | 84.8 | 83.7 |
| COPAR | 85.8 | 85.4 | 84.6 | 83.6 |
| LCLE-DL | 85.2 | 84.3 | 83.5 | 81.3 |
| LRSDL | 86.2 | 85.8 | 85.2 | 84.8 |
| RFDDL-r | **87.1** | **86.8** | **86.4** | **86.2** |
| RFDDL-e | **87.3** | **86.9** | **86.5** | **86.3** |

*I. Hyperparameter Analysis*

We investigate the effects of model parameters $\alpha$, $\beta$ and $\gamma$ on the results of RFDDL-e and RFDDL-r, i.e., the sensitivity of our methods to model parameters. Since the optimal parameter selection still remains an open issue to date, we follow common procedures to use a heuristic way to select the most important parameters. Since there are three parameters, we aim to fix one of them and explore the effects of other two on result by grid search. MIT CBCL face database is used as an example and the number of training samples per class is set to 4. For each pair of parameters, we average the results based on 15 random splits of training/testing sets. We show the parameter selection results of our RFDDL-r and RFDDL-e in Figs. 12 and 13, respectively. We can find that RFDDL-r and RFDDL-e generally perform well in a range of parameters. Specifically, our RFDDL-r with $\alpha > 10^{-2}, \beta > 10^4$ and $\gamma > 10^4$ can generally work well, and our RFDDL-e obtains better results when $\beta > 10^4$. Note that similar observations and findings can be obtained from other databases, but the results are not provided due to the page limitation.

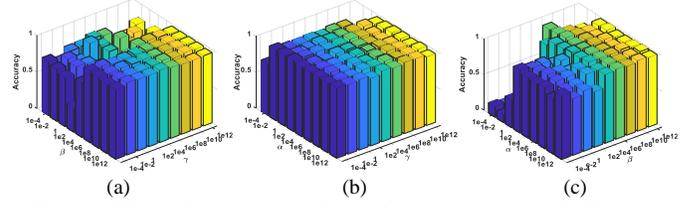

**Fig. 12:** Effects of parameters $\alpha, \beta$ and $\lambda$ on the accuracy of RFDDL-r, where (a) fix $\alpha = 10^2$ to tune $\beta$ and $\gamma$ by grid search. (b) fix $\beta = 10^4$ to tune $\alpha$ and $\gamma$. (c) fix $\gamma = 10^8$ to tune $\alpha$ and $\beta$.

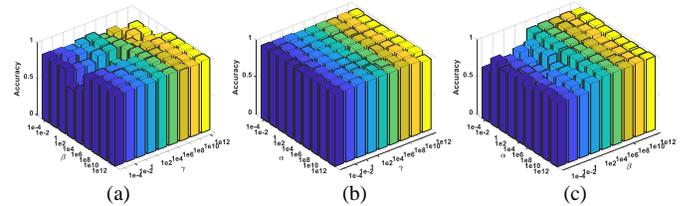

**Fig. 13:** Effects of parameters $\alpha, \beta$ and $\lambda$ on the accuracy of RFDDL-e, where (a) fix $\alpha = 10^2$ to tune $\beta$ and $\gamma$ using grid search. (b) fix $\beta = 10^6$ to tune $\alpha$ and $\gamma$. (c) fix $\gamma = 10^8$ to tune $\alpha$ and $\beta$.

## VI. CONCLUSION AND FUTURE WORK

We have investigated the joint subspace recovery and enhanced discriminative locality driven robust flexible label consistent dictionary learning problem. Our model aims at improving the robustness against noise and outliers in data and dictionary by recovering the clean data and atom subspaces. To encode the representation accurately, our model also proposes the flexible reconstruction error, discriminative flexible sparse code error and flexible classification error, which can enable our model to process the data sampled from a nonlinear manifold potentially, enable the codes to be soft and predict the labels of samples accurately by avoiding the overfitting issue. A discriminative Laplacian matrix is also derived over the recovered atoms to make the neighborhood discriminative and accurate, and by associating it with the learning of codes and classifier jointly to obtain more accurate discriminative codes and classifier.

We mainly examined the effectiveness of our algorithm on public databases. By visualizing the recovered clean data and atoms, we can find that the subspace recovery can remove noise and outliers effectively from the original data and dictionary. Quantitative classification also demonstrate the superiority of our method in terms of performance and robustness. In future, a general strategy for the optimal selection of parameters needs investigation. Although the alternatively updating strategy of RFDDL can converge to a local minimum, the probability that the proposed method converged to the global optimum should be studied theoretically. Besides, because different real-world application data usually deliver different complex distributions and structures, and the inclusion mechanisms of RFDDL-r and RFDDL-e are different, we will also explore the comparison of them in our future work on various kinds of datasets.

ACKNOWLEDGMENT

The authors would like to express sincere thanks to reviewers for their insightful comments, making our manuscript a higher standard. This work is partially supported by National Natural Science Foundation of China (61672365, 61732008, 61725203, 61622305, 61871444 and 61572339), High-Level Talent of the "Six Talent Peak" Project of the Jiangsu Province of China (XYDXX-055), and the Fundamental Research Funds for the Central Universities of China (JZ2019HGPA0102). Dr. Zhao Zhang is the corresponding author of this paper.

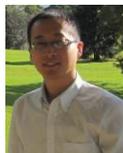

**Zhao Zhang** (SM'17- ) is a Full Professor at the School of Computer Science & School of Artificial Intelligence, Hefei University of Technology, Hefei, China. He received the Ph.D. degree from the City University of Hong Kong in 2013. He was a Visiting Research Engineer at the National University of Singapore, worked with Prof. Shuicheng Yan, from Feb to May 2012. He then visited the Chinese Academy of Sciences (CAS), worked with Prof. Cheng-Lin Liu, from Sep to Dec 2012. During Oct 2013 and Oct 2018, he was an Associate Professor at the Soochow University, Suzhou, China. His current research interests include data mining & machine learning, pattern recognition & image analysis. He has authored/co-authored about 70 technical papers published at prestigious international journals and conferences, including IEEE TIP (3), IEEE TKDE (5), IEEE TNNLS (3), IEEE TCYB, IEEE TSP, IEEE TBD, IEEE TII (2), ACM TIST, Pattern Recognition (6), Neural Networks (5), ICDM, ACM ICMR, ICIP and ICPR, etc. He is now serving as an Associate Editor (AE) of Neurocomputing, IEEE Access and IET Image Processing. Besides, Dr. Zhang is serving/served as a Senior Program Committee (SPC) member of PAKDD 2019/2018/2017, an Area Chair (AC) of BMVC 2018/2016/2015、ICTAI 2018, a PC member for several popular international conferences. He is now a Senior Member of the IEEE.

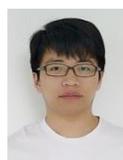

**Jihuan Ren** is currently working toward the research degree at the School of Computer Science and Technology, Soochow University, Suzhou 215006, P. R. China. His current research interests include pattern recognition, machine learning, data mining and their applications. Specifically, he is interested in designing advanced low-rank coding and dictionary learning methods for robust image representation and classification.

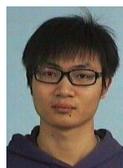

**Weiming Jiang** is now working toward the research degree at the School of Computer Science and Technology, Soochow University, P. R. China. His current research interests include pattern recognition, machine learning and data mining. He has authored or co-authored some papers published in IEEE Trans. on Neural Networks and Learning Systems (IEEE TNNLS), ACM International Conf. on Multimedia Retrieval (ACM ICMR), IEEE Trans. on Industrial Informatics (IEEE TII), and the IEEE International Conf. on Data Mining (ICDM).

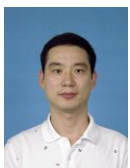

**Zheng Zhang** received the M.S. and Ph.D. degree from the Harbin Institute of Technology in 2014 and 2018, respectively. Currently, he is a Postdoctoral Research Fellow in The University of Queensland, Australia. Dr. Zhang visited the National Laboratory of Pattern Recognition (NLPR) at the Chinese Academy of Sciences (CAS) from June 2015 to June 2016. He was a Research Associate at The Hong Kong Polytechnic University from Apr. 2018 to Oct. 2018. He has authored or co-authored over 40 technical papers published at prestigious international journals and conferences, including the IEEE TPAMI, IEEE TNNLS, IEEE TIP, IEEE TCSVT, Pattern Recognition, CVPR, ECCV, and AAAI, IJCAI, SIGIR, and etc. He received the Best Paper Award from 2014 International Conference on Smart Computing (SMARTCOMP). His current research interests include machine learning and computer vision.

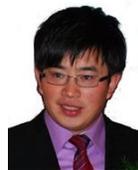

**Richang Hong** (M'12-) received his Ph.D. degrees in July 2008 from the University of Science and Technology of China (USTC). He worked as a research fellow with Prof. Chua Tat-Seng in School of Computing, National University of Singapore (NUS), until Dec. 2010. After that, Dr. Hong joined the School of Computer and Information, Hefei University of Technology (HFUT), as a Professor. He visited Microsoft Research Asia from Mar. to Sept. 2012 by "Star-Track" Program and visited Prof. Qi Tian at University of Texas, San Antonio from Aug. to Nov. 2014. His current research interests include multimedia content analysis and social media. He has authored over 100 journal and conference papers in these areas and the Google Scholar citations for those papers is more than 5000. He served as editor of the IEEE Multimedia Magazine, Information Sciences, Signal Processing and Neural Processing Letter, and the guest editors of several international journals, a technical program chair of the International conference on Multimedia Modeling 2016 and the ACM International Conference on Internet Multimedia Computing and Services 2017. He served as an area chair of ACM Multimedia 2017 and a technical program committee member of over 20 prestigious international conferences, and a reviewer of over 20 prestigious international journals. He is a recipient of the Best Paper Award in ACM Multimedia 2010, Best Paper Award in ACM International Conf. on Multimedia Retrieval 2015 and Best Paper Honorable Mention Award of IEEE trans. Multimedia 2015. Dr. Hong is the CCF technical committee member on multimedia and the secretary of the ACM SIGMM China Chapter.

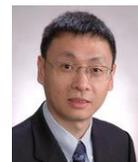

**Shuicheng Yan** (F'16-) received the Ph.D. degree from the Peking University in 2004. He is now the Chief Scientist of Qihoo/360 company, and also a (Dean's Chair) Associate Professor at National University of Singapore. His research areas include computer vision, multimedia and machine learning, and he has authored/co-authored more than 370 technical papers over a wide range of research topics, with the Google Scholar citations >34, 000 times and H-index-73. He is the ISI Highly-cited Researcher from 2014 to 2018. He is/has been an Associate Editor of IEEE Trans. on Knowledge and Data Engineering (TKDE), IEEE Trans. on Circuits and Systems for Video Technology (TCSVT), ACM Transactions on Intelligent Systems and Technology (TIST), and Journal of Computer Vision and Image Understanding. His team received 7 times winner or honorable-mention prizes in PASCAL VOC and ILSVRC competitions, along with more than 10 times best (student) paper prizes. He was the winner of 2010 TCSVT Best Associate Editor (BAE) Award, 2010 Young Faculty Research Award, 2011 Singapore Young Scientist Award and 2012 NUS Young Researcher Award. He is also a Fellow of the IEEE and the IAPR.

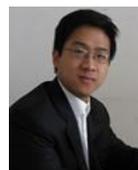

**Meng Wang** is a Full Professor at the Hefei University of Technology, China. He received the B.E. degree and Ph.D. degree in the Special Class for the Gifted Young and signal and information processing from the University of Science and Technology of China, Hefei, China, respectively. His research interests include multimedia content analysis, search, mining, recommendation, and large-scale computing. He has authored 6 book chapters and over 100 journal and conference papers in these areas, including IEEE TMM, TNNLS, TCSVT, TIP, TOMCCAP, ACM MM, WWW, SIGIR, ICDM, etc. He received the paper awards from ACM MM 2009 (Best Paper Award), ACM MM 2010 (Best Paper Award), MMM 2010 (Best Paper Award), ICIMCS 2012 (Best Paper Award), ACM MM 2012 (Best Demo Award), ICDM 2014 (Best Student Paper Award), PCM 2015 (Best Paper Award), SIGIR 2015 (Best Paper Honorable Mention), IEEE TMM 2015 (Best Paper Honorable Mention), and IEEE TMM 2016 (Best Paper Honorable Mention). He is the recipient of ACM SIGMM Rising Star Award 2014. He is/was an Associate Editor of IEEE Trans. on Knowledge and Data Engineering (TKDE), IEEE Trans. on Neural Networks and Learning Systems (TNNLS) and IEEE Transactions on Circuits and Systems for Video Technology (TCSVT). He is a senior member of the IEEE.